\documentclass[sigconf]{acmart}
\copyrightyear{2024}
\acmYear{2024}
\setcopyright{none}
\acmConference[EASE 2024]{28th International Conference on Evaluation and Assessment in Software Engineering}{June 18--21, 2024}{Salerno, Italy}
\acmBooktitle{28th International Conference on Evaluation and Assessment in Software Engineering (EASE 2024), June 18--21, 2024, Salerno, Italy}

\usepackage{graphicx} 
\usepackage{arydshln} 
\usepackage{subcaption}
\usepackage{tablefootnote}

\title{The Promise and Challenges of Using LLMs to Accelerate the Screening Process of Systematic Reviews}

\author{Aleksi Huotala}
\email{aleksi.huotala@helsinki.fi}
\affiliation{%
\institution{University of Helsinki}
\department{Department of Computer Science}
\city{Helsinki}
\country{Finland}}
\authornote{The two first authors contributed equally to this work.}

\author{Miikka Kuutila}
\email{miikka.kuutila@dal.ca}
\affiliation{
\institution{Dalhousie University}
\department{Faculty of Computer Science}
\city{Halifax}
\country{Nova Scotia, Canada}}
\authornotemark[1]

\author{Paul Ralph}
\email{paulralph@dal.ca}
\affiliation{
\institution{Dalhousie University}
\department{Faculty of Computer Science}
\city{Halifax}
\country{Nova Scotia, Canada}}

\author{Mika Mäntylä}
\email{mika.mantyla@{helsinki.fi/oulu.fi}}
\affiliation{%
\institution{University of Helsinki / University of Oulu}
\department{Department of Computer Science / M3S}
\city{Helsinki / Oulu}
\country{Finland}}

\renewcommand{\shortauthors}{Huotala, Kuutila, et al.}

\begin{document}

\begin{abstract}
\textbf{Context:} Systematic review (SR) is a popular research method in software engineering (SE). However, conducting an SR takes an average of 67 weeks. Thus, automating any step of the SR process could  reduce the effort associated with SRs. \textbf{Objective:} Our objective is to investigate the extent to which Large Language Models (LLMs) can accelerate title-abstract screening by (1) simplifying abstracts for human screeners, and (2) automating title-abstract screening entirely. \textbf{Method:} We performed an experiment where human screeners performed title-abstract screening for 20 papers with both original and simplified abstracts from a prior SR. The experiment with human screeners was reproduced by instructing GPT-3.5 and GPT-4 LLMs to perform the same screening tasks. We also studied whether different prompting techniques (Zero-shot (ZS), One-shot (OS), Few-shot (FS), and Few-shot with Chain-of-Thought (FS-CoT) prompting) improve the screening performance of LLMs. Lastly, we studied if redesigning the prompt used in the LLM reproduction of title-abstract screening leads to improved screening performance. \textbf{Results:} Text simplification did not increase the screeners' screening performance, but reduced the time used in screening. Screeners' scientific literacy skills and researcher status predict screening performance. Some LLM and prompt combinations perform as well as human screeners in the screening tasks. Our results indicate that a more recent LLM (GPT-4) is better than its predecessor LLM (GPT-3.5). Additionally, Few-shot and One-shot prompting outperforms Zero-shot prompting. \textbf{Conclusion:} Using LLMs for text simplification in the screening process does not significantly improve human performance. Using LLMs to automate title-abstract screening seems promising, but current LLMs are not significantly more accurate than human screeners. To recommend the use of LLMs in the screening process of SRs, more research is needed. We recommend future SR studies to publish replication packages with screening data to enable more conclusive experimenting with LLM screening.
\end{abstract}


\maketitle

\section{Introduction}

Systematic review (SR) is a popular research method in software engineering; since 1994, over a hundred SRs about software testing~\cite{garousiSystematicLiteratureReview2016} have been published. By a recent estimate, conducting a SR takes about 67 weeks~\cite{marshallSystematicReviewAutomation2019}. A significant amount of time is spent on manual labor, which prior studies have tried to automate ~\cite{vandinterAutomationSystematicLiterature2021, marshallSystematicReviewAutomation2019, guoAutomatedPaperScreening2024, khraishaCanLargeLanguage2024, bolanosArtificialIntelligenceLiterature2024, wilkinsAutomatedTitleAbstract2023, robinsonBioSIEVEExploringInstruction2023,wangZeroshotGenerativeLarge2024}. Our motivation for this study is to investigate whether the effort required for SRs can be reduced by using Large Language Models (LLMs) in text simplification or automating the whole screening process. 

The SR process defines a screening process by which primary studies are included or excluded~\cite{kitchenhamGuidelinesPerformingSystematic2007}. This process usually has two phases, the first using just the title and abstract; and the second using the full text. Title-abstract screening is usually faster and done before full-text screening~\cite{kitchenhamGuidelinesPerformingSystematic2007}. The ability of LLMs to do complex language tasks makes them good candidates for speeding up the screening based on the titles and abstracts~\cite{zhaoSurveyLargeLanguage2023}. Similarly, the ability of LLMs to do text simplification, which has been shown to increase reading comprehension ~\cite{crossley2014s}, suggests their potential for improving screening performance and reducing manual labor in the screening process. We, therefore, pose the following research question: 

\smallskip
\noindent\textit{\textbf{Research Question:} Can Large Language Models (LLMs) facilitate title-abstract screening in systematic reviews (SRs)?}
\smallskip

To address this question, we first perform a controlled experiment comparing the performance of human screeners with original abstracts vs. LLM-simplified abstracts. We hypothesize that text simplification will improve human screening performance by making text more readily understandable. We then evaluate LLM-automated screening by reproducing the screening process of a prior SR using LLMs. We hypothesize that LLMs can do title-abstract screening more accurately than humans and that applying prompt optimization improves screening performance. 

\section{Background}\label{sec:background}

\subsection{Systematic Review Procedures}
SRs are secondary research. \textit{Primary studies} are those in which the authors generate or collect data. \textit{Secondary research} aggregates the findings of primary studies~\cite{ralph2022paving}. Many types of SRs exist, and protocols vary by type; however, SR protocols typically define:

\begin{enumerate}
    \item A search process that identifies primary studies.
    \item Selection criteria that define which primary studies should be included.
    \item A screening process by which selection criteria are applied to classify primary studies as included or excluded~\cite{kitchenhamGuidelinesPerformingSystematic2007, ralph2022paving}
\end{enumerate}

Done well, screening involves two or more human screeners applying selection criteria to (possibly thousands of) primary studies, meeting frequently to resolve disagreements and improve their decision rules. To reduce effort, screening is frequently divided into two stages: 

\begin{enumerate}
    \item \textit{Title and abstract screening:} judges read each paper's title and abstract, excluding those that \textit{unambiguously} do not meet the screening criteria
    \item \textit{Full-text screening:} judges review each paper's complete text, excluding the remaining papers that do not meet the screening criteria.
\end{enumerate}

Overall, conducting SRs can be time- and labor-intensive~\cite{vandinterAutomationSystematicLiterature2021,marshallSystematicReviewAutomation2019}. Any technique that can exclude papers with high accuracy may therefore significantly accelerate the screening process. 

\subsection{Text Simplification}
Text simplification is used to improve the readability and understandability of text while retaining its original meaning~\cite{al2021automated}. Text simplification is often used with second language learners, children, people with disabilities, and people with low literacy~\cite{vstajner2021automatic}. It seems plausible that text simplification could improve abstract comprehension by human screeners, especially for non-native English speakers. 

\subsection{Scientific Literacy}
\label{sec:scientific_literacy}
The ability to correctly screen primary studies depends on the screener's scientific literacy. \textit{Scientific literacy} is the ability to ``use evidence and data to evaluate the quality of science information and arguments put forth by scientists and in the media'' \cite{gormally2012developing}. Put another way, it is ``the capacity to use scientific knowledge to identify questions and to draw evidence-based conclusions in order to understand and help make decisions about the natural world and the changes made to it through human activity''~\cite{gormally2012developing}. Scientific literacy skills thus concern the use of data, evaluating the quality of scientific literature, making decisions, and reaching conclusions. Scientific literacy can be evaluated using a standardized test such as the Test Of Scientific Literacy Skills (TOSLS)~\cite{gormally2012developing}.

\subsection{Using Software to Help Systematic Reviews}
\label{sec:software_to_help_systematic_reviews}

Software should be able to automate or accelerate several aspects of SRs including constructing search strings, retrieving PDFs, extracting data, and automating screening. However, most existing studies focus on automating primary study screening and acknowledge that the main goal of automation is to reduce manual labor~\cite{vandinterAutomationSystematicLiterature2021}. Prior work on using software to facilitate SRs focuses on reducing manual effort but stops short of automating the whole review process. For example, a review of text mining techniques used on SRs found that text mining techniques could reduce effort by 30--70\%, but led to a $\approx 5\%$ loss of relevant studies~\cite{o2015using}. 

Marshall and Wallace~\cite{marshallSystematicReviewAutomation2019} provide a practical guide for using machine learning (ML) tools for literature search, screening, data extraction, or bias assessment. Several abstract-screening tools are available, of which \textsc{Abstrackr} and \textsc{Research} \textsc{Screener} tools reduce screening workload by 9--89\%, with precision ranging from 16\% to 45\%~\cite{chaiResearchScreenerMachine2021,gatesTechnologyassistedTitleAbstract2018, rathboneFasterTitleAbstract2015a}. All of these tools require a human in the loop, instructing the ML tool to identify relevant research papers. 

Recent studies on using LLMs to automate the screening process in SRs have shown that LLMs can match human performance in paper screening, with some limitations~\cite{guoAutomatedPaperScreening2024, khraishaCanLargeLanguage2024, bolanosArtificialIntelligenceLiterature2024, wilkinsAutomatedTitleAbstract2023, robinsonBioSIEVEExploringInstruction2023,wangZeroshotGenerativeLarge2024}. However, most of these approaches have been studied in the domain of medical sciences, the findings suggest that research on automating title-abstract screening could also be applied to other domains, such as SE.

\subsection{Prompt Optimization}
\label{sec:learning_strategies}
Incorporating task-specific examples and specially crafted instructions into LLM prompts has been shown to improve their performance in specific tasks~\cite{weiChainofThoughtPromptingElicits2023, kojimaLargeLanguageModels2023}. Zero-shot (ZS) prompting refers to situations where no task-specific examples are provided. One-shot (OS) and Few-shot (FS) prompting are typical prompting techniques in natural language processing (NLP), providing task-specific examples~\cite{kojimaLargeLanguageModels2023}. OS prompting incorporates one task example, while FS prompting includes multiple examples, with the number of examples depending on the context size of the LLM~\cite{brownLanguageModelsAre2020}. Chain-of-Thought (CoT) prompting incorporates a ``Let's think step by step'' phrase at the end of the prompt, instructing the LLM to decompose the task into smaller sub-tasks for more accurate task solving~\cite{weiChainofThoughtPromptingElicits2023}. CoT prompting can be combined with FS prompting, which is known as Few-shot with Chain-of-Thought (FS-CoT) prompting, to further improve task accuracy~\cite{weiChainofThoughtPromptingElicits2023}.
\section{Methodology} \label{sec:methodology}

\begin{figure*}
\resizebox{\textwidth}{!}{%
\includegraphics{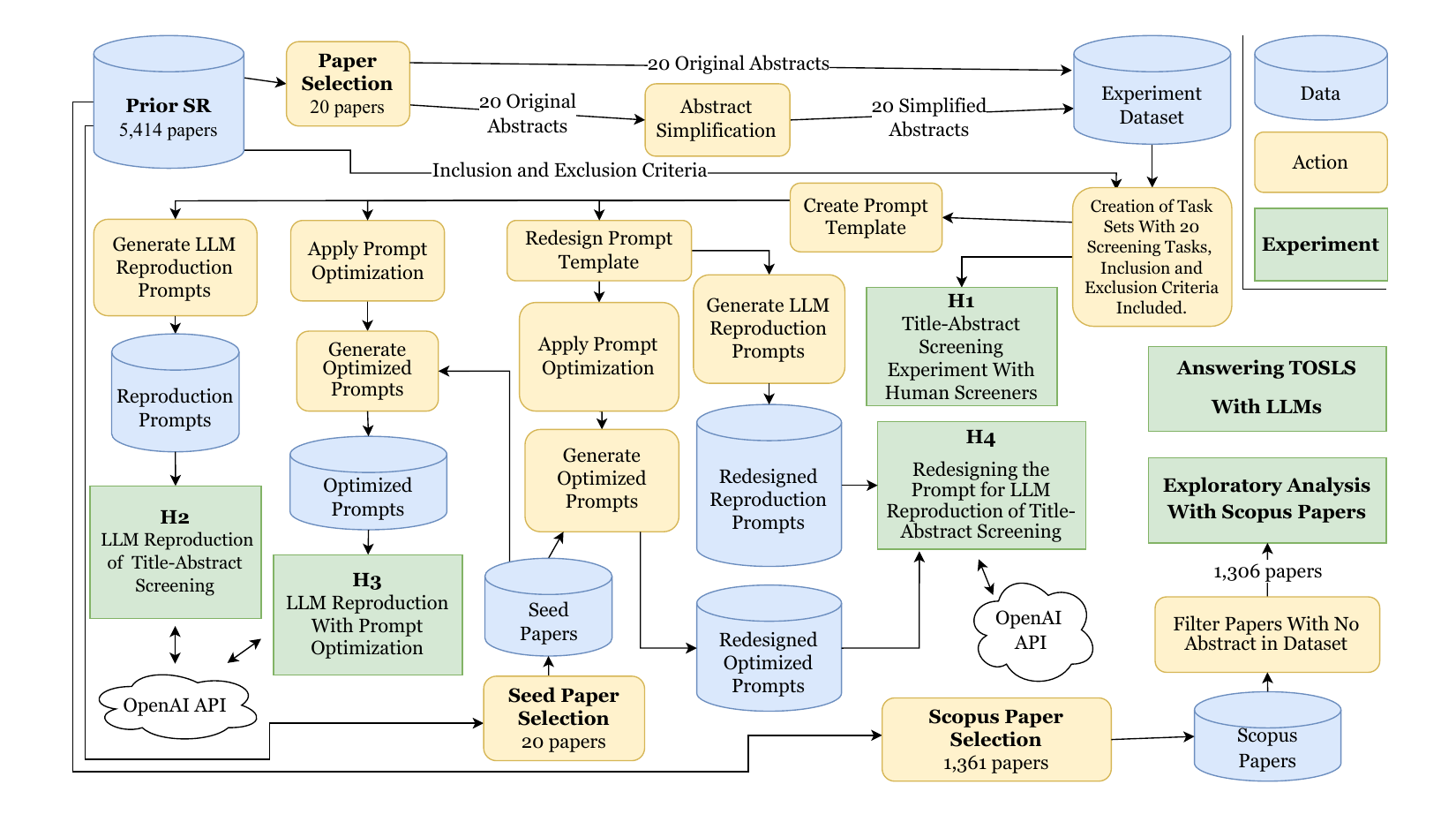}
}%
\Description[Diagram detailing the conducted experiments, datasets and actions.]{Diagram detailing the conducted experiments, datasets and actions.}
\caption{Diagram detailing the conducted experiments, datasets and actions.}
\label{fig:experiment_setup}
\end{figure*}

Figure \ref{fig:experiment_setup} provides a visual overview of our study. Our study consists of two experiments related to title-abstract screening for SRs. In the first experiment, we investigate the effect of abstract simplification by LLMs on the performance of humans in title-abstract screening. In the second experiment, we examine the efficacy of completely automated title-abstract screening using different LLMs and prompting techniques. We refer to this second experiment as the LLM reproduction experiment because it aims to reproduce the first experiment but with screening decisions made by LLMs. We provide a comprehensive replication package including data, source code, and materials (see \nameref{sec:data_availability}).

\subsection{Hypotheses}
For the purposes of this study, we formulated the following hypotheses:

\def\firsthypothesis{Using LLMs to simplify abstracts improves human screening performance}
\def\secondhypothesis{LLMs screen articles better than human screeners}
\def\thirdhypothesis{LLM selection and prompt optimization improve the screening performance}
\def\fourthhypothesis{Redesigning the LLM reproduction prompt leads to improved screening performance}

\renewcommand{\theenumi} {H\arabic{enumi}}
\begin{enumerate}
    \item \firsthypothesis
    \item \secondhypothesis
    \item \thirdhypothesis
    \item \fourthhypothesis
\end{enumerate}
\renewcommand{\theenumi}{\arabic{enumi}}

Hypothesis (H1) concerns the experiment with human screeners. It was formulated on the basis that a simplified abstract is easier to comprehend, enhancing the screening performance of screeners.

Hypothesis (H2) suggests that the tested LLMs outperform humans in the screening task. We believe that LLMs can potentially identify key concepts better and also detect irrelevant information better than humans, leading to improved screening performance. 

Hypothesis (H3) suggests that LLM selection and prompt optimization improve screening performance, compared to the LLM reproduction in (H2). As noted in Section \ref{sec:learning_strategies}, prompt optimization has been shown to improve task accuracy in other contexts. Thus as prior work supports prompt optimization to increase task accuracy, we assert that prompt optimization can increase the LLM's performance in title-abstract screening. We also posit that LLM selection can improve the performance in screening because newer LLMs are trained with more data and contain improved architectures.

Finally, hypothesis (H4) suggests that redesigning the prompt, used in hypothesis (H2) (the LLM reproduction), leads to improved screening performance. The rationale for hypothesis (H4) is the usage of a prompt, that 1) describes the inclusion and exclusion criteria in more detail; and 2) is formulated in a way that the LLM can make more accurate decisions when including and excluding articles.

\subsection{Abstract Simplification}\label{methdodsimplification}

We conducted an experiment to study the effects of text simplification, scientific literacy, and academic experience on the screening performance of human screeners. 

From a pool of 5,414 papers originally screened for a prior SR~\cite{kuutila_time_2020}, we selected 10 papers that were eventually included and 10 that were eventually excluded (top-left of Figure~\ref{fig:experiment_setup}). We used our expertise to purposefully select difficult-to-screen papers. All selected papers have time-related terminology in their abstracts and mention empirical results related to time pressure. In the excluded papers, time pressure is used as a motivating factor for building something (e.g., testing framework), or time-related terminology is used in relation to other concepts such as temporal dispersion. 

The inclusion criteria in the prior SR were:
    \textit{(1) ``The main focus is time pressure in software engineering``; and (2) ``The paper presents empirical evidence of time pressure in software engineering``.}~\cite{kuutila_time_2020}

We selected papers from this particular SR because two of the authors of this paper were co-authors, the subject matter is relatively easy for non-experts to understand, and the search process included many synonyms. 

\textit{Abstract simplification} (top middle in Figure \ref{fig:experiment_setup}) was done using ChatGPT version 3.5. The prompt used to simplify abstracts was: 
\begin{quote}
    \textit{Simplify the following text so that a high school student can understand it, mention \textbf{X} on the simplified text and make the simplified text at least \textbf{Y} words long.}
\end{quote} 

\noindent where \textbf{X} was the identified time-related term of each original abstract and \textbf{Y} was a number slightly less than the number of words in the original abstract. The prompt was used as many times as necessary to get an abstract that included the time-related terms identified in the original abstract. 

Table \ref{tab:abstracts} presents the lengths of the original and simplified abstracts and two readability metrics. The readability metrics were calculated using version 3.3.1 of the Quanteda library for the R programming language~\cite{benoit2018}. The Flesch-Kincaid Grade level (FK)~\cite{kincaid1975derivation} and Simple Measure of Gobbledygook (SMOG)~\cite{mc1969smog} are readability metrics that estimate the US educational grade level (FK) or years of education (SMOG) required to completely understand a text. That is, lower scores indicate easier-to-read text. All selected papers and their original and simplified abstracts can be found in the replication package (see \nameref{sec:data_availability}). 

\begin{table}[t]
\centering
\caption{Original vs. simplified abstracts in the experiment with human screeners.}  
\label{tab:abstracts}
\begin{tabular}{llll}
Means & Sample & Original & Simplified \\
\hline
Abstract length (words) & include & 228.9 & 196.6 \\
Abstract length (words) & exclude & 181.4 & 176.7\\
\hdashline
Flesch-Kincaid level & include & 16.99 & 11.33 \\
Flesch-Kincaid level & exclude & 15.43 & 10.79 \\
\hdashline
SMOG & include & 17.48 & 12.78 \\
SMOG & exclude & 16.18 & 12.50 \\
\hdashline
Accuracy & & 69\% & 69\%\\
Time per task (seconds) & & 85.95 & 76.26\\
Confidence & &  7.77 & 7.79 \\
\hline
\end{tabular}
\end{table}

\subsection{Title-Abstract Screening Experiment With Human Screeners} \label{experiment_with_human_raters}

\begin{table}[!htb]
\caption{Demographic information, test of scientific literacy (TOSLS) scores and screening performance of humans by group.}  
\label{tab:participants}
\begin{tabular}{l|ll}
\hline
     & MSc Students & Researchers \\
     
\hline
Respondents &  13 & 16 \\
\hdashline
Male & 11 & 8\\
Female & 2 & 8\\
\hdashline
Nationalities* & Finnish (8) & Finnish (8)\\
& Chinese (2) & Iran (2)\\
& Portuguese (2) & Others (8)\\
& Other (1) & \\
\hdashline
Min Age & 21 & 23\\
Max Age & 33 & 64\\
Mean Age & 26.2 & 39.8 \\
\hdashline
Mean Academic Exp & 0 & 4.3 \\
Mean Industry Exp & 1 & 6.5 \\
\hdashline
Mean TOSLS & 19 & 18.9\\
\% correct tasks & 0.64 & 0.73\\
Avg time for a task & 82.77s & 79.94s\\
Avg confidence for a task & 7.08 & 8.34\\
\hline
\end{tabular}
\flushleft*Dual-citizens marked for both nationalities.

\end{table}

We recruited both master's students and researchers at a Finnish university. The students were enrolled in either a combined bachelor's / master's degree program, or an international master's degree program, both of which are  course-based --- students are not expected or paid to produce publishable research. In contrast, all of the researchers were paid to do research and publish their results. The student screeners participated in the experiment voluntarily as part of a course, where participation netted extra points for completion. Researcher participants were recruited through social contacts and mailing lists and were compensated with items of their choosing from a university gift shop.

The experiment began with a presentation on what the screeners would be doing. Participants were told they could ask questions about the experiment at any time. Participants then signed a consent form and completed a demographic questionnaire including age, nationality, gender, and highest academic qualification (see Table ~\ref{tab:participants}). Next, participants had 35 minutes to complete the TOSLS (see Section~\ref{sec:scientific_literacy}). 

Each paper screening task consisted of an A4 paper, where the start time of screening would be marked by hour, minute, and second. The reverse side presented first the title and then the abstract of the paper. This was followed by the inclusion and exclusion rules, where the participant's choice was crossed on a checkbox. After this, the confidence in the judgment was asked with the single item ``On a scale of 1 to 10, how confident are you that your previous answer is correct?''. Lastly, the time the paper was completed was marked at the bottom of the page. The invigilator demonstrated the filling of the task sheet, first from the time started and then turning the paper, and so on. The invigilator emphasized that there was no time limit for this task. For time recording purposes, a digital clock with seconds was visible to participants at all times during this part. 

Each participant was given a set of four sets of five studies to screen: 
\begin{enumerate}
    \item original abstracted / excluded from original study
    \item original abstracted / included in original study
    \item simplified abstracted / excluded from original study
    \item simplified abstracted / included in original study
\end{enumerate}

The creation of the task sets is the last action before the H1 experiment in Figure \ref{fig:experiment_setup}. Before creating the task sets, we divided the 40 possible tasks into two creation sets with 20 tasks, where the conditions for each paper were mirrored, that is the same paper had the original abstract in the first set and the simplified one in the second. Both of these creation sets had 5 tasks with the aforementioned 4 conditions. By doing this we ensured that each condition for each paper got a similar amount of ratings. The task sheets were then printed out and it was made so that the first four tasks were one of each of the four conditions. Additionally, these were made so that roughly quarter of the participants received each condition as their first task and so on, for each of the first four tasks. After the fourth task, the remaining tasks were shuffled.

At the end, there was a debriefing about the purposes of the study, where the aims and methodology of the study were briefly explained, and the participants could ask further questions if they so wished.

\subsection{LLM Reproduction of Title-Abstract Screening}  \label{inclusion_exclusion_direct_replication}

We reproduced the experiment from Section \ref{experiment_with_human_raters} using LLMs, as shown in Figure \ref{fig:experiment_setup}. We used the \textit{gpt-3.5-turbo-0613} and \textit{gpt-4-0613} chat model implementations, as GPT-3.5 and GPT-4 are commonly used LLMs in SE-related research \cite{houLargeLanguageModels2023}. The corresponding LLM configurations are shown in Table \ref{tab:llm_configuration}. As the prompts in this experiment fit in the context of both models, the 16K context version (\textit{gpt-3.5-turbo-16k-0613}) is unnecessary. We used a temperature value of zero to minimize the non-deterministic behavior of the tested LLMs, although some variation remains visible. The model cut-off date for all LLMs was set to \textit{June 2023} and a seed parameter of \textit{128} was used in this and consequent experiments to make comparing the results more coherent and reproducible.

The reproduction uses the same screening task set of 20 papers from the prior SR ~\cite{kuutila_time_2020}, as in the experiment done with humans. We constructed a prompt template following the screening procedure of the prior SR as closely as possible. Using this template, we generated prompts using the ZS prompting technique (see Section \ref{sec:learning_strategies}) to use in the reproduction. These prompts were then processed with a Python script, which called the OpenAI API. The prompt template, the generated prompts, the Python script, and the API responses are provided in the replication package (see \nameref{sec:data_availability}).

\begin{table}[t]
    \centering
\caption{LLM configuration parameters.}
\resizebox{\linewidth}{!}{%
    \begin{tabular}{l|llll}
        Model                     & t  & Seed & Context & Cut-off date \\
        \hline
        gpt-3.5-turbo-0613        & 0.0          & 128 & 4K & June 13th 2023  \\
        gpt-3.5-turbo-16k-0613    & 0.0          & 128 & 16K & June 13th 2023 \\
        gpt-4-turbo-0613          & 0.0          & 128 & 8K & June 13th 2023 \\
    \end{tabular}
}%
    \label{tab:llm_configuration}
\end{table}

\subsection{LLM Reproduction With Prompt Optimization}
\label{prompt_optimization_experiment}

The experiment for applying prompt optimization to the reproduction prompts roughly follows the design of the simple reproduction experiment, as detailed in Section \ref{inclusion_exclusion_direct_replication}. There are, however, some changes in the prompt generation process, to accommodate the requirements of prompt optimization.

We selected 20 \textit{additional} papers from the prior SR~\cite{kuutila_time_2020} as ``seed papers'', of which 10 should be included and 10 should be excluded. The seed papers were used as examples with OS, FS, and FS-CoT prompting (see Section \ref{sec:learning_strategies}). 

For OS prompting, we generated 20 prompts for each of the 20 seed papers, producing a table with 20 rows and 20 columns for each seed paper and their corresponding screening result. For FS prompting, we used a subset of the seed papers from the OS prompting. We specifically selected four seed papers that performed the best in the OS prompting experiment; two each with the highest $F_{1}$-scores from the included and excluded categories. We used seed papers from the GPT-3.5 OS prompting experiment to maintain model specificity, and likewise for GPT-4.

For FS-CoT prompting, we reused the FS prompts. The phrase ``Let's think step by step'' was appended to the end of each prompt. This phrase is used in Chain-of-Thought (CoT) prompting and improves task accuracy \cite{weiChainofThoughtPromptingElicits2023, kojimaLargeLanguageModels2023}.

\subsection{Redesigning the Prompt for LLM Reproduction of Title-Abstract Screening}
\label{initial_prompt_redesign}

The prompt redesign experiment closely follows the LLM reproduction and the LLM reproduction with prompt optimization experiments from Sections \ref{inclusion_exclusion_direct_replication} and \ref{prompt_optimization_experiment}, respectively. 

To redesign the prompt, we first chose two title-abstract pairs to run our evaluations against. We chose pairs that were challenging for both humans and LLMs to evaluate. Second, we instructed ChatGPT (GPT-3.5) to optimize the prompt using the prompt template from the LLM reproduction experiment. If the optimization led to an incorrect screening decision, we ran the task again. Once the evaluation was successful, we manually removed any information from the prompt that was irrelevant for making the screening decision based on the provided title-abstract pairs. However, after the removal of non-redundant information, we observed that the screening decision was again incorrect. To solve this, we appended ``Even shallow empirical evidence should be considered.'' to the end of the inclusion criteria.

The redesigned prompt is included in the replication package (see \nameref{sec:data_availability}), which shows that: 1) the list of synonyms was streamlined into the first inclusion criterion to improve readability and broaden the scope of aspects related to time pressure in software engineering; and 2) the second inclusion criterion was made more inclusive, to allow even shallow empirical evidence, as long as it is related to the impact of time-related factors on software engineering. Together, these two modifications make the prompt shorter, essentially easier to understand by humans and LLMs.

For the experiments incorporating ZS and OS prompting, the remaining test setup is identical to the prompt optimization setup described in Section \ref{prompt_optimization_experiment}, using the optimized prompt template (see \nameref{sec:data_availability}). Furthermore, we did not test FS and FS-CoT prompting and decided to solely focus on experimenting with ZS and OS prompting.

\subsection{Answering TOSLS With LLMs} \label{tosls_direct_replication}

To answer the TOSLS from Section \ref{experiment_with_human_raters} with LLMs, a prompt using ZS prompting was constructed (included in the replication package; see \nameref{sec:data_availability}). The prompt includes the task description and the 28 questions of the TOSLS. Nine of the questions contain background information that incorporates images. Out of the nine images, six were manually converted to a textual representation. For the remaining three images (questions 10, 21, and 28), which are challenging for a human to describe textually, we used ChatGPT's (GPT-4) image interpretation functionality. The images were also described textually so we could answer the whole TOSLS with LLMs that don't support embedding images. For question 10, we used the instruction ``Describe the following website for me.'' to convert the website image to a textual representation. For questions 21 and 28, the instruction used was ``Describe the graph below for me.''.

The experiment was conducted using the \textit{gpt-3.5-turbo-16k-0613} and \textit{gpt-4-0613} chat model implementations, as described in Table \ref{tab:llm_configuration}. The TOSLS in prompt form has an approximate token count of 5700, so we have to use the 16K context version of the \textit{gpt-3.5-turbo-0613} chat model. For the two LLMs, the TOSLS was run 10 times in succession. We report the average result of the 10 test runs.

\subsection{Reproducing a Larger Screening Procedure} \label{replicating_procedure}

We also conducted exploratory analysis on 1,361 Scopus Papers that were initially gathered for the prior SR ~\cite{kuutila_time_2020} to see how LLMs perform with a more realistic sample size (see Figure \ref{fig:experiment_setup}). The Scopus Papers do not encompass all the screened articles for the prior SR, as the screened articles also included those from Google Scholar and articles resulting from snowballing, but it is the data that had complete abstracts readily available for analysis. Articles without abstracts or to which we did not have access were removed, leaving us with 1,306 papers.

From the Scopus Papers, we extracted the papers' corresponding titles and abstracts. For the title-abstract pairs, we constructed prompts using the ZS prompting technique (see Section \ref{sec:learning_strategies}) and then followed the same procedure used for the LLM reproduction (Section \ref{inclusion_exclusion_direct_replication}).

\subsection{Statistical Analysis} \label{statistical_analysis}

To analyze the data from all of the steps above, we used a variety of statistical procedures. We checked for normality with the Shapiro-Wilk test. The Shapiro-Wilk test was significant at p = 0.045 for variables of total number of correct screening tasks by rater for original abstracts (value between 0 and 10). Thus, we used the non-parametric paired Wilcoxon signed rank test for hypothesis H1. The Shapiro-Wilk test was not significant for the total number of correct screening tasks by researchers and students, with p-values of 0.33 and 0.59 respectively. Thus we used the Independent samples t-test when comparing the total number of correct screening tasks (value between 0 and 20) between students and researchers for exploratory analysis.

For the Ordinary Least Squares (OLS) regression model presented in Table~\ref{tab:linear_regression}, the predicted variable is the number of correct screening tasks from the twenty given to the participant, and the predictor variables are the total TOSLS score and a categorical variable on whether the screener was a researcher or a student.

To compare human screener performance with LLMs for Hypothesis H2, we generated a dataset where each LLM and prompt combination was presented with 29 sets of tasks similar to human screeners, thus producing 580 ratings. Thus we had the same amount of data for human screeners and each LLM and prompt combination. The sets also mirrored those given to human screeners in other characteristics, that is each set contained 10 inclusions, 10 exclusions, 10 original abstracts, and 10 simplified abstracts.

This dataset was then used to fit the logistic regression model presented in Table~\ref{tab:logistic_regression_combinations}, where the predicted variable is the correctness of a single judgment, and it is predicted by a categorical variable capturing who (human or LLM \& prompt combination) made the judgment. We also added control variables to the model on whether the abstract was simplified or the original, and whether the paper in the task was included or excluded in the original paper. We chose human performance as the reference category because of the way our hypothesis was formulated.

For testing hypotheses H3 and H4, we recategorized the categorical variables from the LLM and prompt combination into two: the LLM used and the prompt used. We also took out the judgments by human screeners as they are not relevant to the hypotheses, and they would have a perfect correlation if used for both LLM and prompt variables. The model is presented in Table~\ref{tab:logistic_regression_llmprompts}. We chose GPT-3.5 and ZS prompt as reference categories, as they are the most basic versions which the subsequent versions try to improve on. We give Odd's ratio as an effect size measure for both logistic regression models and $R^2$ for the OLS linear regression.
\section{Results}\label{sec:results}

\subsection{Text Simplification}
\subsubsection{H1: \firsthypothesis}

\textbf{The statistical analysis did not support hypothesis H1}. Table \ref{tab:wilcoxon} presents Wilcoxon Signed-Rank tests comparing original abstracts with simplified abstracts. Based on the results, it is evident that whether the abstract was original or simplified did not improve the performance of the human screeners in screening tasks or the confidence in their screening. However, screeners spent less time screening simplified abstracts, as can be seen in Table \ref{tab:wilcoxon}. As can be seen from Table \ref{tab:abstracts}, the difference between the means is 9.39 seconds (76.26s vs 85.95s).

\begin{table}[t]
\caption{Wilcoxon Signed-Rank Test (Paired) between correctness, time and confidence between original and simplified paper abstracts} 
\label{tab:wilcoxon}
\begin{tabular}{l|l}
Comparison  & p-value \\
\hline
Original v Simplified correctness & 0.61 \\
Original v Simplified time &  0.0007*** \\
Original v Simplified confidence  & 0.75 \\
\hline
\end{tabular}
\end{table}

\subsubsection{Exploratory Analysis}
Researchers exhibit an average correctness of 73\% in paper screening, while students screen the paper correctly 64\% of the time. Based on the Independent samples t-test, the difference between these groups is statistically significant, as shown in Table \ref{tab:ttest}. There was no meaningful difference in TOSLS correctness between students and researchers, as shown in Table \ref{tab:participants}.

Linear regression model predicting the total number of correct judgments for human screeners is shown in Table \ref{tab:linear_regression}, utilizing student/researcher status and TOSLS score as predictors. Our findings indicate that both a higher TOSLS score and researcher status are predictive of an increased number of accurate screening decisions. This model accounts for 33\% of the variance in screening decisions.

\begin{figure}[t]
\Description[Percentage of correct screening tasks and TOSLS score scatter plot with Spearman correlation.]{Percentage of correct screening tasks and TOSLS score scatter plot with Spearman correlation.}
\centering
\includegraphics[width=\linewidth]{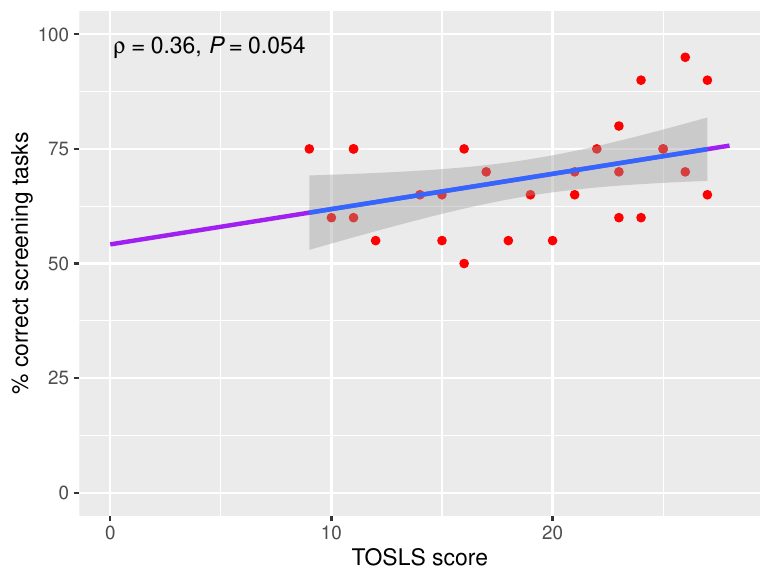}
\caption{Percentage of correct screening tasks and TOSLS score scatter plot with Spearman correlation.}
\label{fig:totaltoslsscatter}
\end{figure}

Figure \ref{fig:totaltoslsscatter} displays a Spearman correlation of 0.36 and a scatter plot between the TOSLS score and the total number of correct screening tasks. It is noteworthy that the regression line intersects the y-axis at nearly 50\% correctness, aligning with the expected outcome from random guessing in a binary classification task. Furthermore, the regression model presented in Table \ref{tab:linear_regression} indicates that the TOSLS score is a statistically significant positive predictor of correct evaluations. Thus, scientific literacy, the TOSLS score in particular, has a predictive ability on the correctness of the screening tasks.

\begin{table}[t]
\caption{Independent samples t-test between the total number of correct screening tasks by students and researchers.}  
\label{tab:ttest}
\begin{tabular}{l|l|l}
Comparison & t-value & p-value \\
\hline
Students v Researchers & 2.4227 & 0.02277 \\
\hline
\end{tabular}
\end{table}

\begin{table}[t]
\caption{OLS linear regression model predicting total number correct screening tasks by human screeners.} 
\label{tab:linear_regression}
\begin{tabular}{l|lllll}
Variable & coef & std.err & t & p-value & 95\% CI \\
\hline
 TOSLS & 0.17 & 0.06 & 2.48 &0.02* & 0.03 0.29\\
 Student & -1.81 & 0.71 & -2.56 &0.02* & -3.27 -0.35\\
\hline
$R^2$:0.33 & \multicolumn{5}{l}{F-Stat >0.001***} \\
\hline
\end{tabular}
\end{table}

\newcommand{\specialcell}[2][c]{%
  \begin{tabular}[#1]{@{}c@{}}#2\end{tabular}}

\begin{table*}[!htb]
	\caption{Human screener majority vote per paper. Paper references are given in the replication package ~\cite{zenodo}.}
	 
	\resizebox{\linewidth}{!}{%
		\begin{tabular}{c|cccccccccc:cccccccccc}
			Paper                                  & A1   & A2   & A3   & A4   & A5   & A6   & A7  & A8   & A9   & A10                                    & A11  & A12  & A13  & A14  & A15  & A16  & A17  & A18  & A19  & A20           \\
			\hline
			\specialcell{Include count \\Original}                          & 2    & 3    & 4    & 2    & 4    & 5    & 0   & 8    & 1    & 3                                      & 3    & 4    & 9    & 10   & 14   & 13   & 6    & 9    & 13   & 6             \\
			\hdashline
			\specialcell{Exclude count \\Original}                          & 13   & 11   & 10   & 12   & 10   & 10   & 14  & 8    & 14   & 12                                     & 12   & 10   & 5    & 3    & 1    & 2    & 9    & 5    & 1    & 9             \\
			\hdashline
			\specialcell{Include count \\Simplified}                          & 10   & 1    & 2    & 8    & 4    & 5    & 0   & 5    & 5    & 0                                      & 4    & 5    & 10   & 14   & 14   & 12   & 9    & 11   & 11   & 6             \\ 
			\hdashline
			\specialcell{Exclude count \\Simplified}                          & 4    & 14   & 13   & 7    & 11   & 9    & 15  & 8    & 9    & 14                                     & 10   & 10   & 5    & 2    & 0    & 2    & 5    & 4    & 4    & 8             \\ 
			\hdashline
			\% Correct Orig                        & 86.6 & 78.6 & 71.4 & 85.7 & 71.4 & 66.6 & 100 & 50   & 93.3 & 80                                     & 20   & 28.6 & 64.3 & 76.9 & 93.3 & 86.6 & 40   & 64.3 & 92.8 & 40            \\
			\% Correct Simp                        & 28.6 & 93.3 & 86.6 & 46.6 & 73.3 & 64.3 & 100 & 61.5 & 64.3 & 100                                    & 28.6 & 33.3 & 66.6 & 87.5 & 100  & 85.7 & 64.3 & 73.3 & 73.3 & 42.9          \\
			\hdashline
			Total \% Correct                       & 59   & 86   & 79   & 66   & 72   & 66   & 100 & 55   & 69   & 90                                     & 24   & 31   & 66   & 83   & 97   & 86   & 52   & 69   & 83   & 41            \\
			\hdashline
			Judgment in Orig \cite{kuutila_time_2020} & \multicolumn{10}{c}{Exclude} & \multicolumn{10}{|c}{Include} \\
		\end{tabular}
	}%
	\label{tab:include_exclude_results}
\end{table*}

\begin{table}[!htb]
    \caption{Mean TOSLS scores of the two tested LLM models.}
    \centering
    
\resizebox{\linewidth}{!}{%
    \begin{tabular}{l|c|c|c}
      Name   &  Mean TOSLS  & $\rho$ & p \\
    \hline
      Human screeners & 19.068 & & \\
      \hdashline
      gpt-3.5-turbo-16k-0613  & 19.000 & 0.54115 & 0.10623 \\
      gpt-4-0613    &  28.000 & 0.89352 & 0.00049 \\
    \end{tabular}
    }%
    \label{tab:tosls_llm_results}
\end{table}

\subsection{LLM Screening}

\begin{table}
	\centering
	\caption{Results of the title-abstract screening of the reproduction of larger screening procedure, using Scopus data set.}
	\label{tab:larger_screening_procedure_results}
	\begin{tabular}{l|l|l|l|l}
		Correct (\%) Model       & Exclude & Include & Precision & \textbf{Recall}  \\
		\hline
		gpt-3.5-turbo-0613 & 95.25\% & 65.00\% & 0.65      & \textbf{0.17567} \\
		gpt-4-0613         & 98.91\% & 50.00\% & 0.5       & \textbf{0.41667} \\
	\end{tabular}
\end{table}

\begin{table}[!htb]
\caption{Logistic regression model predicting the correctness of a single judgment with the combination of LLM and prompting technique. Reference category is the category the following categories are compared to.} 
\label{tab:logistic_regression_combinations}
\resizebox{\linewidth}{!}{%
\begin{tabular}{l|lrrrr}
LLM \& prompt & coef & std.err & t & p-value & \specialcell{Odd's\\ratio}   \\
\hline
Reference: & \multicolumn{5}{{l}}{Human performance} \\
GPT3.5 - FS     &       -0.56 &    0.12 &  -4.56 & <0.001 & 0.57 \\
GPT3.5 - FS-CoT  &       -0.48 &    0.12 &  -3.86 & <0.001 & 0.62 \\
GPT3.5 - OS    &        -0.42 &    0.12&  -3.38 & <0.001 & 0.66 \\
GPT3.5 - OSv2  &        -0.43 &    0.12&  -3.44 & <0.001 & 0.65 \\
GPT3.5 - ZS    &        -0.91 &    0.12&  -7.36 & <0.001 & 0.40 \\
GPT3.5 - ZSv2  &        -0.26 &    0.13 &  -2.12 & 0.03 &  0.77 \\
GPT4 - FS      &        0.02 &    0.13 &   0.19 & 0.85 &  1.02  \\
GPT4 - FS-CoT   &        0.13 &    0.13 &   1.03 & 0.30 & 1.14  \\ 
GPT4 - OS      &        0.02 &    0.13 &   0.13 & 0.89 & 1.02   \\ 
GPT4 - OSv2    &       -0.63&    0.12 &  -5.08& <0.001 & 0.53 \\
GPT4 - ZS      &       -0.20&    0.13 &  -1.63 & 0.10 & 0.82  \\
GPT4 - ZSv2   &        -0.31 &    0.12 &  -2.48 & 0.01 & 0.73 \\
\hdashline
Reference: & \multicolumn{5}{{l}}{Exclude} \\
Include  &  0.40  &   0.05  &   8.39 & <0.001 & 1.50 \\
\hdashline
Reference: & \multicolumn{5}{{l}}{Original abs} \\
Simplified abstract & -0.08 & 0.06  & -1.61& 0.11 & 0.93\\
\hline
\end{tabular}
}%
\footnotesize{ \raggedright FS = Few-shot, FS-CoT = Few-shot with Chain-of-Thought, OS = One-shot, OSv2 = redesigned One-shot, ZS = Zero-shot, ZSv2 = redesigned Zero-shot. Refer to Section \ref{sec:learning_strategies} for more in-depth explanations about the prompting techniques. \\
}
\end{table}

\begin{table}[!htb]
\caption{Logistic regression model predicting the correctness of a single judgment with the LLM and prompting technique as their own variables. Reference category is the category the following categories are compared to.} 
\label{tab:logistic_regression_llmprompts}
\resizebox{\linewidth}{!}{%
\begin{tabular}{l|lllll}
Variable & coef & std.err & t  & p-value & \specialcell{Odd's\\ratio}  \\
\hline
Reference: LLM - GPT-3.5  \\
LLM - GPT-4      &    0.34 &   0.05 & 6.90 & <0.001 & 1.41 \\
\hdashline
Reference: Prompt - ZS \\
Prompt - FS      &     0.29  &  0.09 &  3.33 & <0.001& 1.33  \\
Prompt - FS-CoT   &     0.38  &  0.09 &  4.41 & <0.001 & 1.46 \\
Prompt - OS      &      0.36  &  0.09 &  4.15 & <0.001 & 1.43 \\
Prompt - OSv2    &      0.04  &  0.08 &  0.42 & 0.67 &   1.03   \\
Prompt - ZSv2    &      0.28  &  0.09 &  3.24 & 0.0012 &  1.32  \\
\hdashline
Reference: Exclude \\
Include & 0.47  &  0.05 &  9.49 & <0.001 & 1.61 \\
\hdashline
Reference: Original abs\\
Simplified abstract &  -0.08 & 0.05 & -1.64 & 0.10 & 0.92 \\
\hline
\end{tabular}
}
\footnotesize{ \raggedright FS = Few-shot, FS-CoT = Few-shot with Chain-of-Thought, OS = One-shot, OSv2 = redesigned One-shot, ZS = Zero-shot, ZSv2 = redesigned Zero-shot. Refer to Section \ref{sec:learning_strategies} for more in-depth explanations about the prompting techniques. \\

}
\end{table}

\subsubsection{H2: \secondhypothesis}

As can be seen in Table \ref{tab:logistic_regression_combinations}, none of the LLM and prompt combinations have both a positive coefficient and a statistically significant p-value. Thus \textbf{hypothesis H2 is rejected}. However, there are several LLM and prompt combinations that perform roughly as well as human screeners, specifically these are One-shot (OS), Few-shot (FS) and Few-shot with Chain-of-Thought (FS-CoT) prompts with the GPT-4 LLM.

The worst performance is by the Zero-shot (ZS) prompt for GPT-3.5 LLM. This not too surprising, as the zero shot prompt is the most rudimentary of the prompts. The next worst combinations in order are OSv2 for GPT-4, and FS, FS-CoT, OSv2, OS for GPT-3.5. Lastly the ZSv2 for both GPT-3.5 and GPT-4 both perform also worse than humans, but not as significantly.  

\subsubsection{H3: \thirdhypothesis}

The logistic regression model in Table \ref{tab:logistic_regression_llmprompts} shows that using prompt optimization improves performance over ZS prompts. Table also shows that GPT-4 is superior to GPT-3.5  \textbf{Thus, based on these results, we accept hypothesis H3}. All prompt optimization techniques, that is OS, FS, and FS-CoT prompting, improve screening performance over ZS prompting, with odds ratios of 1.33 to 1.46. This means that all of these prompts are between 33\% to 46\% more likely to make a correct judgment on a single task. GPT-4 is 41\% more likely to make a correct judgment than GPT-3.5.

\subsubsection{H4: \fourthhypothesis}

The regression model in Table \ref{tab:logistic_regression_llmprompts} shows that redesigning the prompt (marked with v2 text in the table) improves screening performance for ZS prompting, having a 32\% increased chance of making a correct judgment on a single task. However, OS prompting with the redesigned prompt shows no statistical significance. \textbf{Based on these observations, we reject hypothesis H4}.

\subsection{Exploratory Analysis}

We did an exploratory analysis on conducting the TOSLS with LLMs and reproducing the title-abstract screening on a larger screening procedure. Table \ref{tab:tosls_llm_results} shows the mean TOSLS score for humans and both tested LLMs and Table \ref{tab:larger_screening_procedure_results} presents the results of the larger title-abstract screening procedure.

In the TOSLS experiment, the mean score of humans is very close to the score of \textit{gpt-3.5-turbo-16k-0613}, but \textit{gpt-4-0613} scored full points from the TOSLS. For the Scopus paper screening experiment, both GPT-3.5 and GPT-4 classify correctly over 95\% of excluded papers. However, this comes with a significant tradeoff, missing 35 to 50\% of included papers for GPT-3.5 and GPT-4, respectively. Therefore, using Zero-shot prompts for the whole screening procedure cannot be recommended.

In the TOSLS experiment involving LLMs, we suspect that GPT-4 may have been trained with the TOSLS research paper, so it might be aware of the test questions and their answers. Unfortunately, as OpenAI does not disclose specifics about the training data for their models, we cannot be certain. The mean TOSLS for students was 19 and for researchers 18.9, which means GPT-3.5 performed as well as students and researchers. The skill scores of GPT-3.5 are weakly correlated to human scores, meanwhile, GPT-4 indicates a very strong correlation. If GPT-4 did not know the correct answers to the TOSLS, the full points from the test would indicate significantly improved reasoning abilities.
\section{Discussion}\label{sec:discussion}

The experiment with the human screeners revealed no significant benefits on accuracy when using summarized abstracts with GPT-3.5. However, simplified abstracts resulted in significantly faster completion of screening judgments. On average, the screening task for simplified abstracts was made in 9.39 seconds, or around 12\% faster. This demonstrates that our text simplification procedure had a real and measurable impact. However, considering that in a real-life use case, researchers would have to generate the simplified versions of the abstracts themselves, the actualized benefits would be limited. However, even if this benefit was significant enough, the results in Table \ref{tab:include_exclude_results} call for caution, as it seems that the simplified text was much more likely included erroneously in some cases (see papers A1, A4 and A9).  

Paid researchers outperformed students in the screening task. This result is partly in conflict with the results of the experiment conducted by Salman et al.~\cite{salman2015students}, where it was concluded that student performance was similar to professional performance in a programming task when the approach used was novel for both groups. In our case, the students had been taught about conducting secondary studies during the course they were attending, thus it was not completely novel for them, we did not record the experience of the researchers in performing them. Nevertheless, a noticeable result from the experiment with human screeners is that being a researcher and having higher scientific literacy skills, measured with the TOSLS, has a predictive ability on the number of correct judgments according to our regression model. 

In the second experiment (H2), neither GPT-3.5 nor GPT-4 screened papers better than humans. However, GPT-4 performed significantly better than GPT-3.5, and we expect this performance improvement to continue with newer versions of LLMs, as the amount of training data increases and architectures are improved.

Out of the tested prompting techniques, we observed that FS-CoT prompting performed the best when used with the GPT-4 LLM. Additionally, we observed that using seed papers included in the prior SR resulted in increased performance, that is prompts incorporating OS or FS prompting outperformed ZS prompts. In our experiment setup, we used an additional set of 20 papers as seed papers for the prompts and picked the best-performing seed papers. Thus, for increased performance of these prompts are to be actualized in conducting actual SRs, the authors have to have a set of screened papers so that best performing seed papers can be evaluated. Thus prompt optimization would be easier to use when conducting SRs on topics with prior SRs, as their screening results could be used for One-shot (OS) and Few-shot (FS) prompting. This underscores the importance of sharing the screening results in published SRs. 

The results for redesigning the prompt were mixed at best. The ZSv2 prompt outperformed ZS with both GPT-3.5 and GPT-4 LLMs, but the OSv2 prompt did not outperform OS with either LLMs. It is worth noting that the redesigned prompts (ZSv2, OSv2) were generally outperformed by prompt optimization techniques (OS, FS, FS-CoT). Thus it seems that prompt redesign techniques have limited performance increase in the screening task, and concentrating on prompt optimization is more warranted. Future work could still examine if complex prompting optimization strategies, such as FS and FS-CoT, could perform better together with the redesigned prompt.  

Further work is needed to analyze prompt optimization methods on larger datasets. Our ZS prompt, which was fairly simple, already achieved a 98.91\% accuracy in excluding papers correctly in the exploratory analysis. It would make sense to sacrifice some of this accuracy to increase inclusion accuracy, as the needed manual labor for screening the included papers is already significantly reduced if a significant amount of papers can be automatically excluded. Thus, investigating how to safely integrate the usage of LLMs in the screening process while keeping integrity of the existing guidelines for SRs is called for.

The two existing ML tools, \textsc{Abstrackr} and \textsc{Research Screener} (covered in Section~\ref{sec:software_to_help_systematic_reviews}) had a screening accuracy between 16\% and 45\%~\cite{chaiResearchScreenerMachine2021,gatesTechnologyassistedTitleAbstract2018, rathboneFasterTitleAbstract2015a}. Compared to our exploratory testing on a larger screening procedure, in Table~\ref{tab:larger_screening_procedure_results}, our screening performance seems significantly higher than with \textsc{Abstrackr} and \textsc{Research Screener}. However, the results are not directly comparable, as the datasets used were different, and \textsc{Abstrackr} gives recommendations instead of judgments. Also of note is that we picked 20 screening tasks for our experiments, and the performance for these particularly difficult cases is between 50\% and 75\% with a recall between 0.3 to 0.9. Precision and recall for each LLM and prompt combination is given in the replication package~\cite{zenodo}. The recent studies on LLMs for automating the screening process of SRs~\cite{guoAutomatedPaperScreening2024, khraishaCanLargeLanguage2024, bolanosArtificialIntelligenceLiterature2024, wilkinsAutomatedTitleAbstract2023, robinsonBioSIEVEExploringInstruction2023,wangZeroshotGenerativeLarge2024} show that the accuracy of title-abstract screening with LLMs have varied between 0.60 and 0.835, which are similar to our results. Similar to the studies done with \textsc{Abstrackr} and \textsc{Research Screener}, our results are not directly comparable with the recent studies on LLMs, because of the datasets that focus on different aspects. Additionally, not all studies mention the used LLM versions, seed or temperature parameters making comparisons harder.

\section{Limitations}
\label{sec:limitations}
One concern of validity is the accuracy of the original study's judgments regarding screening. It is worth noting that the majority vote by the human screeners in determining screening judgment aligns with the original study 17 out of 20 times. Yet, this was exceeded by the three best individuals that made correct decisions 19, 18 and 18 times respectively. The three instances where there is a discrepancy between the original study and the majority vote, the human screeners of this study exclude studies, even though they were included in the original review paper. These abstracts were particularly challenging in judgment, but we maintain that each should be included, following the screening criteria of the original study.

Another concern is the generalizability of our results. We selected particularly difficult papers to screen for our experiment, where as the whole screening procedure for a systematic review can involve very straightforward decisions depending on the criteria. We reckon that both human screeners and LLMs would have had better performance with a complete set of papers from an existing study.

Using a digital clock could have added pressure for the participants to complete the experiment faster. However, the researcher overseeing the experiment emphasized to the participants that the clock was there for recording the times of the individual screening tasks, and there was no need to hurry to complete the tasks.

In our experiments with LLMs, we used date-fixed versions of GPT-3.5 and GPT-4 LLMs. It is evident that these models will get outdated in the future, but we wanted to make sure the experiments can be exactly replicated, if necessary. Another limitation regarding LLMs is the exclusive use of GPT-3.5 and GPT-4 as the LLMs in our experiments. The choice was intentional to focus on the most commonly used LLMs in SE-related research. However, the emergence of new LLMs, like Google Gemini 1.5 and Anthropic Claude 3 pave possibilities for future research on additional LLMs.

As conducting SRs includes more than just paper screening, another avenue for future work is considering the use of LLMs in more phases. This includes the identification of studies for the screening~\cite{wohlin2022successful}, data extraction and analysis.
\section{Conclusions}\label{sec:conclusions}

In this paper, we studied the effects of text simplification, scientific literacy, and academic experience in the screening accuracy of human screeners for an SR and the accuracy of LLMs in reproducing the screening task of a prior SR. We also studied if applying prompt optimization in the LLM reproduction of title-abstract screening improves screening performance, compared to the base case.

Our findings indicate that text simplification does not affect the number of correct screening decisions, but it does reduce the time spent on paper screening. Additionally, our linear regression model reveals that higher scientific literacy skills and being a researcher (as opposed to a student) both contribute to increased accuracy in screening results.

None of the LLM and prompt combinations outperformed human screeners in the reproduction experiment. Nevertheless, Few-shot, Few-shot Chain-of-Thought (CoT), and One-shot prompts together with GPT-4 LLM performed roughly as well as the human screeners. This shows that there is promise in automating title-abstract screening with LLMs. However, we note that the comparison here is to a sample consisting of master's level students and researchers who were not experts on the specific subject. The reproduction experiment also shows that prompt optimization techniques improve the screening performance, whereas the results with prompt redesign were mixed. Thus based on our results the use of prompt optimization is recommended. Still, further investigation on integrating the LLM automated title-abstract screening to the SR process is needed before it can be recommended for practical application.

\section*{Data Availability}
\label{sec:data_availability}

The datasets and source code for the experiments are available at Zenodo ~\cite{zenodo}. However, the data involving human screeners is not shared due to the absence of explicit consent for distribution.

\begin{acks}
\label{sec:acknowledgements}

The first author has been funded by the Strategic Research Council of Research Council of Finland (Grant ID 358471) and the second author has been funded by the Killam Postdoctoral Fellowship.

\end{acks}

\bibliographystyle{ACM-Reference-Format}
\bibliography{sources}

\end{document}